\documentclass[runningheads]{llncs}

\usepackage{eccv}

\usepackage{eccvabbrv}

\usepackage{graphicx}
\usepackage{booktabs}

\usepackage[accsupp]{axessibility}  

\usepackage{hyperref}

\usepackage{orcidlink}
\usepackage[dvipsnames]{xcolor}
\usepackage{booktabs,multirow,array}
\usepackage[T1]{fontenc}
\usepackage{amsmath}
\usepackage{bm}
\usepackage{physics}
\usepackage{subcaption}
\usepackage{silence}
\definecolor{airforceblue}{rgb}{.3,.3,.8}
\newcommand{\airforceblue}[1]{{\color{airforceblue}#1}}

\newcommand{\rv}[1]{{\boldsymbol{\mathbf{#1}}}}
\newcommand{\vaeenc}{q_{\rv{\psi}}(\rv{z} \vert \rv{X})}
\newcommand{\vaedec}{p_{\rv{\xi}}(\rv{X} \vert \rv{z})}
\newcommand{\ldm}{p_\rv{\zeta}(\rv{z} \vert \rv{z}_T)}
\newcommand{\vaeprior}{p(\rv{z})}
\newcommand{\normaldist}[2]{\mathcal{N}\left( #1, #2 \right)}
\newcommand{\unitsphere}{S^2}

\let\titleold\title
\renewcommand{\title}[1]{\titleold{#1}\newcommand{\thetitle}{#1}}
\def\maketitlesupplementary
   {
   \newpage
       {
        \centering
        \Large
        \textbf{\thetitle}\\
        \vspace{0.5em}Supplementary Material \\
        \vspace{1.0em}
       } 
   }

\AtEndPreamble{
    \usepackage[capitalize]{cleveref}
    \crefname{section}{Sec.}{Secs.}
    \Crefname{section}{Section}{Sections}
    \Crefname{table}{Table}{Tables}
    \crefname{table}{Tab.}{Tabs.}
}

\RequirePackage{enumitem}
\setlist[itemize]{noitemsep,leftmargin=*,topsep=0em}
\setlist[enumerate]{noitemsep,leftmargin=*,topsep=0em}

\makeatletter
\DeclareRobustCommand\onedot{\futurelet\@let@token\@onedot}
\def\@onedot{\ifx\@let@token.\else.\null\fi\xspace}

\def\ie{\emph{i.e}\onedot} 
 
 \def\vs{\emph{vs}\onedot}

\makeatother

\begin{document}

\title{FrePolad: \airforceblue{F}requency-\airforceblue{Re}ctified \airforceblue{Po}int \airforceblue{La}tent \airforceblue{D}iffusion for Point Cloud Generation} 

\titlerunning{FrePolad}

\author{Chenliang Zhou\inst{1}\orcidlink{0009-0001-1096-1927} \and Fangcheng Zhong\inst{1}\orcidlink{0000-0001-5964-5282}, Param Hanji\inst{1}\orcidlink{0000-0002-7985-4177}, Zhilin Guo\inst{1}\orcidlink{0000-0002-7660-3102}, Kyle Fogarty\inst{1}\orcidlink{0000-0002-1888-4006}, Alejandro Sztrajman\inst{1}\orcidlink{0000-0002-3312-3272}, Hongyun Gao\inst{1}\orcidlink{0000-0001-5546-6906}, Cengiz Oztireli\inst{1}\orcidlink{0000-0002-4700-2236}}

\authorrunning{C.~Zhou et al.}

\institute{Department of Computer Science and Technology,\\
Cambridge, University of Cambridge, UK}

\maketitle


\begin{abstract}
We propose \emph{FrePolad: \textbf{f}requency-\textbf{re}ctified \textbf{po}int \textbf{la}tent \textbf{d}iffusion}, a point cloud generation pipeline integrating a variational autoencoder (VAE) with a denoising diffusion probabilistic model (DDPM) for the latent distribution. 
FrePolad simultaneously achieves high quality, diversity, and flexibility
in point cloud cardinality
for generation tasks while maintaining high computational efficiency.
The improvement in generation quality and diversity is achieved through (1) a novel frequency rectification via spherical harmonics designed to retain high-frequency content while learning the point cloud distribution; and (2) a latent DDPM to learn the regularized yet complex latent distribution. In addition,
FrePolad supports variable point cloud cardinality by formulating the sampling of points as conditional distributions over a latent shape distribution. Finally, the low-dimensional latent space encoded by the VAE contributes to FrePolad's fast and scalable sampling.
Our quantitative and qualitative results demonstrate FrePolad's state-of-the-art performance in terms of quality, diversity, and computational efficiency. 
Project page: \url{https://chenliang-zhou.github.io/FrePolad/}.
\end{abstract} 
\begin{figure}[t]
    \centering
    \includegraphics[width=0.63\textwidth]{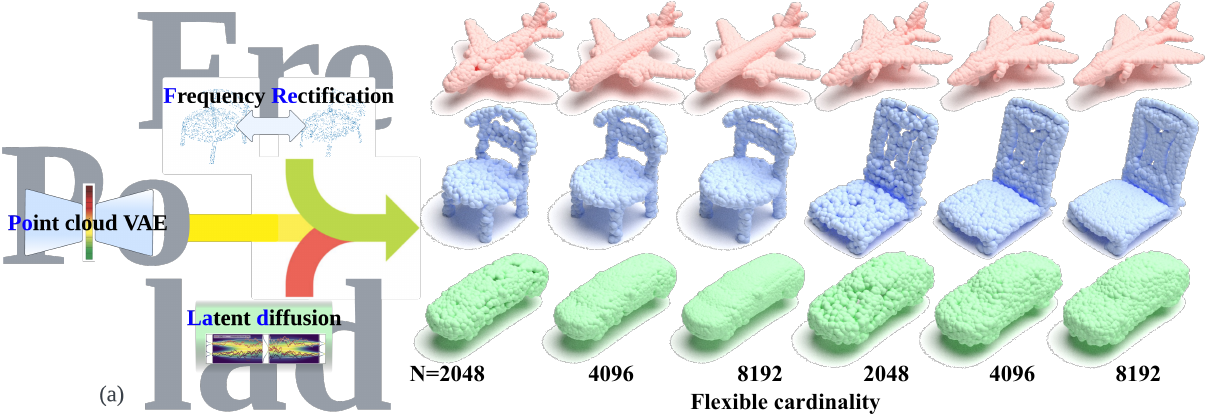}
    \includegraphics[width=0.32\textwidth]{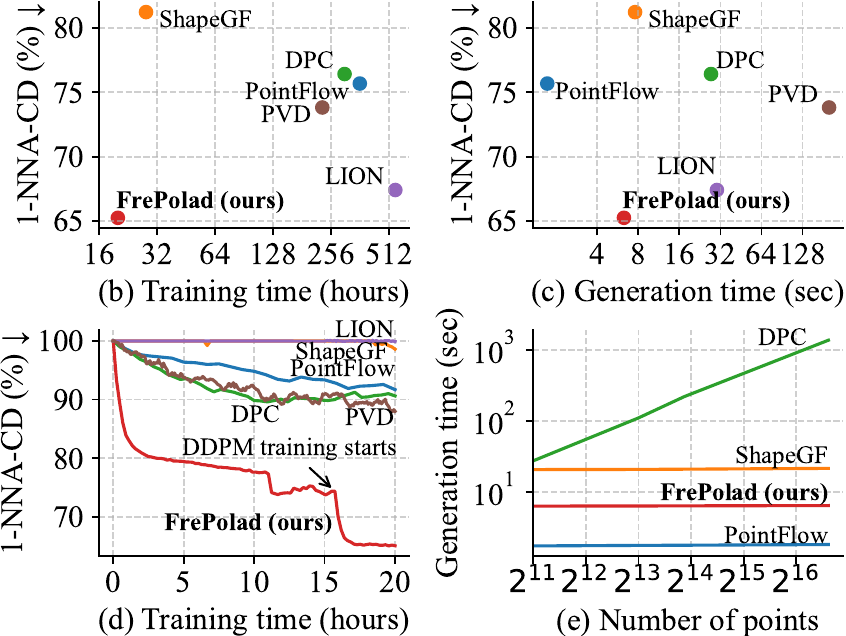}
    \caption{(a) FrePolad combines novel frequency rectification with a point cloud VAE and a DDPM-based prior to generate point clouds with superior quality, diversity, and flexibility in cardinality. Plots show on the right (b) training and (c) generation costs \vs final validation score measured by 1-NNA-CD ($\downarrow$), (d) learning curves for the first 20 hours of training, and (e) generation cost for synthesizing different numbers of points} 
    \label{fig:teaser}
\end{figure}
\section{Introduction}
\label{sec:intro}

Point clouds are a widely-used 3D representation with extensive applications in 3D content creation and reconstruction~\cite{gadelha2018multiresolution, kim2023pointinverter, wu2023fast, yu2018ec, kim2023pointinverter, zhou20213d, fan2017point, kurenkov2018deformnet, groueix2018papier}.
Automated 3D content creation using point clouds has been extensively explored in recent studies through the adoption of generative methods, such as
autoencoders~\cite{yang2018foldingnet, kim2021setvae}, adversarial training~\cite{zamorski2020adversarial, kim2023pointinverter}, flow-based models~\cite{klokov2020discrete, kim2020softflow}, and denoising diffusion probabilistic models (DDPM)~\cite{luo2021diffusion, vahdat2022lion}.
However, addressing the intricate task of learning the distributions of point clouds while simultaneously attaining exceptional sample \emph{quality}, \emph{diversity}, and \emph{computational efficiency} poses a substantial challenge.
For example, variational autoencoders (VAEs)~\cite{kingma2013auto} and generative adversarial networks (GANs)~\cite{goodfellow2014generative} tend to produce noisy or blurry outcomes~\cite{dosovitskiy2016generating, esser2021taming, johnson2016perceptual}, possibly attributed to the \emph{loss of high-frequency information}~\cite{dosovitskiy2016generating, esser2021taming, lee2022autoregressive} caused by the \emph{spectral bias}~\cite{kim2022zoom, mildenhall2021nerf} (also see \cref{sec:related-work-Loss of high frequency in VAE}). Excessive emphasis on generation fidelity, such as through the deployment of excessively deep networks, can lead to reduced diversity and even mode collapse~\cite{li2021hsgan, zhang2022pointot}. VAEs also struggle with the \emph{prior hole problem}~\cite{vahdat2021score, tomczak2018vae, rosca2018distribution} when employing oversimplified Gaussian priors~\cite{chen2016variational}. While normalizing flows~\cite{rezende2015variational} and DDPMs~\cite{sohl2015deep, ho2020denoising} have emerged as powerful generative models, they face challenges in learning high-dimensional distributions, particularly for point clouds. A key issue with both classes of models is the necessity for the latent distribution to match the dimensionality of the target. Thus, despite their exceptional sample quality and diversity in various domains~\cite{kingma2018glow,whang2021composing,mustafa2022distilling,dhariwal2021diffusion,song2021solving,poole2022dreamfusion}, directly applying these models to point cloud generation results in only moderate success~\cite{zhou20213d, luo2021diffusion, yang2019pointflow}.

Generating point clouds with flexible cardinality (varying number of points) is also a critical, yet often overlooked feature in recent works.
Prior works with fixed cardinalities~\cite{gadelha2018multiresolution, achlioptas2018learning} often fail to incorporate the inherent property of permutation invariance in point clouds.
The methods that are trained with heuristic loss functions such as Chamfer distance (CD) or earth mover’s distance (EMD) can also distort the probabilistic interpretation of VAEs~\cite{yang2019pointflow}. 
Furthermore, the flexibility in cardinality is crucial for tasks involving non-uniform datasets, such as those from LiDAR scans~\cite{geiger2013vision, serna2014paris}. Networks with unrestricted cardinality readily scale for tasks with unspecified output point count requirements, avoiding the significant computational overhead associated with retraining fixed-cardinality models, while demonstrating superior quality compared to post-processing the point cloud with upsampling (see \cref{sec:upsampling}).

To simultaneously address the outlined challenges, we propose a novel generative framework, \emph{FrePolad: \textbf{f}requency-\textbf{re}ctified \textbf{po}int \textbf{la}tent \textbf{d}iffusion}. FrePolad integrates a point cloud VAE with a latent DDPM modeling its latent distribution. Given that VAEs typically have access to a low-dimensional latent space, leveraging DDPMs to learn the VAE-encoded latent distributions enhances complex distribution modeling~\cite{pandey2022diffusevae, vahdat2022lion}, preserves high-frequency contents~\cite{vahdat2021score, rombach2022high}, and reduces computational demands (\cref{sec:exp-computational-efficiency}). Furthermore, harnessing insights from spherical harmonic analysis, we introduce a novel \emph{frequency rectification} technique for training the point cloud VAE to further strengthen the preservation of high-frequency contents, thereby significantly boosting the sample fidelity and diversity (\cref{sec:ablation-study}). Finally, 
FrePolad supports a variable number of points in the training set via a permutation-invariant set encoder~\cite{qi2017pointnet, qi2017pointnet++}. 
By formulating the sampling of points as a conditional distribution over a latent shape distribution, \textit{i.e.}, a \emph{distribution of distributions}, FrePolad enables arbitrary cardinality in point cloud generation.
Our extensive evaluation of FrePolad on the ShapeNet dataset~\cite{chang2015shapenet} demonstrates its state-of-the-art performance in terms of generation quality and diversity, while incurring the least computational expense (\cref{fig:teaser}).


Our contributions are summarized as follows:
\begin{itemize}
    \item A novel point cloud generative network realized through a variational autoencoder (VAE) with a prior distribution based on a denoising diffusion probabilistic model (DDPM) (\cref{sec:formulation});

    \item An novel \emph{frequency rectification} technique designed to promote the preservation of high-frequency details during VAE training (\cref{sec:formulation-Frequency Extraction via Spherical Harmonics,sec:formulation-Frequency-Rectified VAE});

    \item A comprehensive demonstration of FrePolad's state-of-the-art performance in terms of quality, diversity, and computational efficiency through extensive experiments and analysis (\cref{sec:experiments}).
\end{itemize}
\section{Related works}
\paragraph{Denoising diffusion probabilistic models}
\label{sec:related-works-Denoising diffusion probabilistic models}

Recent advancements in denoising diffusion probabilistic models (DDPMs)~\cite{sohl2015deep, ho2020denoising} have demonstrated remarkable performance in various domains, including image synthesis~\cite{dhariwal2021diffusion, ho2020denoising, rombach2022high, ramesh2022hierarchical, brooks2023instructpix2pix}, density estimation~\cite{kingma2021variational}, and speech synthesis~\cite{chen2020wavegrad, jeong2021diff, liu2022diffgan}.
Nonetheless, DDPMs are known to lack a low-dimensional, interpretable space~\cite{pandey2022diffusevae} and to be time-inefficient during generation. Given that variational autoencoders (VAEs) typically have access to a low-dimensional latent space,
researchers have explored the potential of DDPMs to model VAE-encoded latent distributions across various modalities, including images~\cite{ramesh2022hierarchical, sinha2021d2c}, music~\cite{mittal2021symbolic}, videos~\cite{he2022latent, zhou2022magicvideo, blattmann2023align}, and point cloud generation~\cite{yang2019pointflow, vahdat2022lion}. Along this direction, our work further probes the efficacy of this paired framework in the domain of point cloud generation.

\paragraph{Point cloud generation}
\label{sec:related-work-Point cloud generation}
A multitude of studies have delved into point cloud generation within various generative frameworks, including auto-regressive autoencoding~\cite{kingma2013auto, rezende2014stochastic, yang2018foldingnet, gadelha2018multiresolution, kim2021setvae}, adversarial generation~\cite{goodfellow2014generative, zamorski2020adversarial, shu20193d, li2018point, achlioptas2018learning, liu2019point, kim2023pointinverter}, flow-based models~\cite{dinh2014nice, kingma2018glow, yang2019pointflow, klokov2020discrete, kim2020softflow}, and, recently, DDPMs~\cite{zhou20213d, wu2023fast, luo2021diffusion, vahdat2022lion, nakayama2023difffacto, wu2023fast, mo2023dit, liu2019point}. Recently, LION~\cite{vahdat2022lion} proposes the VAE framework based on PVCNNs~\cite{liu2019point} with a hierarchical latent space modeled by two DDPMs. Nevertheless, while a hierarchical latent space can be effective in capturing complex distributions~\cite{ramesh2022hierarchical, ho2022cascaded, preechakul2022diffusion}, an overly intricate network topology and multilayered latent codes -- where a single layer exceeds the original point cloud's dimensionality -- may lead to challenges such as overfitting~\cite{piotrowski2013comparison}, lack of interpretability~\cite{zhang2021survey}, and excessive computational cost during both training and sampling.

On the other hand, most works in point cloud generation primarily target at modeling the data distribution of point clouds within $\mathbb{R}^{3 \times N}$ with a fixed number $N$ of points~\cite{gadelha2018multiresolution, achlioptas2018learning}. The limitation of such approach has been discussed in \cref{sec:intro} and in many other works~\cite{luo2021diffusion, yang2019pointflow}.
In contrast, contemporary research has explored a different approach conceptualizing point clouds as probabilistic distributions in $\mathbb{R}^3$, leading to the modeling of a \emph{distribution of distributions}~\cite{yang2019pointflow, luo2021diffusion, sun2020pointgrow, klokov2020discrete, yang2018foldingnet, cai2020learning}.

\paragraph{Frequency analysis in VAE}
\label{sec:related-work-Loss of high frequency in VAE}
Variational autoencoders (VAEs)~\cite{kingma2013auto} are probabilistic generative networks modeling the probability distributions of datasets. A pervasive challenge associated with VAEs is their tendency to attenuate high-frequency data~\cite{dzanic2020fourier}, often resulting in blurry or low-resolution samples~\cite{dosovitskiy2016generating, hou2017deep, esser2021taming, johnson2016perceptual, lee2022autoregressive}. This degradation becomes even more pronounced at higher compression rates~\cite{lin2023catch}. Such loss of high-frequency details can be attributed to the intrinsic \emph{spectral bias} in general neural network layers~\cite{kim2022zoom, rahaman2019spectral, mildenhall2021nerf, tancik2020fourier} and/or to the upsampling modules in VAEs~\cite{chandrasegaran2021closer, wang2021frequency, durall2020watch} during reconstruction or generation.

Various VAE architectures have been proposed to promote data reconstruction in the frequency domain~\cite{gao2021neural, jiang2021focal, xu2019frequency}. Predominantly, these architectures utilize Fourier features~\cite{rahimi2007random} or positional encoding~\cite{vaswani2017attention, mildenhall2021nerf} to preserve high-frequency details~\cite{jiang2021focal, lin2023catch, czolbe2020loss, gao2021neural}. Yet, a limitation exists: they lack generalizability to point cloud data since Fourier analysis requires data with a well-defined grid structure -- such as 2D grids in images or 3D grids in voxels. Drawing inspiration from~\cite{naderi2023lpf}, in \cref{sec:formulation-Frequency Extraction via Spherical Harmonics,sec:formulation-Frequency-Rectified VAE}, we adopt spherical harmonic analysis for frequency information extraction from point clouds.
\section{Background}
\subsection{Variational Autoencoder}
\label{sec:background-vae}
We use a VAE~\cite{kingma2013auto} as our latent distribution model as it provides access to a low-dimensional latent space and has been successfully applied to generate point clouds~\cite{li2022editvae, wang2020learning, li2022spa}. VAEs are probabilistic generative models that can model a probability distribution of a given dataset $\mathcal{X}$.
A VAE consists of an encoder $\vaeenc$ and a decoder $\vaedec$ parametrized by $\rv{\psi}$ and $\rv{\xi}$, respectively. Assuming that the latent codes $\rv{z} \in \mathbb{R}^{D_\rv{z}}$ follow a prior distribution $\vaeprior$, the encoder and decoder are jointly trained to maximize the \emph{evidence lowerbound (ELBO)}:
\begin{equation}
    \begin{aligned}
\label{eq:vae-elbo-no-beta}
    \mathcal{L}_\text{ELBO} (\rv{\psi}, \rv{\xi}; \rv{X}) := \mathbb{E}_{\vaeenc} \left[ \log \vaedec \right] - \mathcal{D}_\text{KL} \left( \vaeenc, \vaeprior \right),
\end{aligned}
\end{equation}
where $\mathcal{D}_\text{KL}$ is the Kullback-Leibler divergence between the two distributions~\cite{csiszar1975divergence}.


\subsection{Denoising Diffusion Probabilistic Model}
\label{sec:background-ddpm}
Our generative framework leverages a denoising diffusion probabilistic model (DDPM)~\cite{sohl2015deep, ho2020denoising} to model the complexity of the VAE latent space. Given a data sample $\rv{z} \sim p(\rv{z})$, at each time step $t=1,2,\ldots,T$, DDPMs gradually transform $\rv{z} =\rv{z}_0$ into $\rv{z}_T$ by adding noise in a Markovian diffusion process according to a predetermined variance schedule $\{\beta_t\}_t$. If $T$ is sufficiently large (1000 steps in practice), $p(\rv{z}_T)$ will approach the standard Gaussian distribution $\normaldist{0}{I}$. Furthermore, since all transition kernels of the diffusion process are Gaussian: $p(\rv{z}_t \vert \rv{z}_{t-1}) \sim \mathcal{N}(\rv{z}_t; \sqrt{1 - \beta_t} \rv{z_{t-1}}, \beta_t I), \forall t$, samples from the intermediate distributions can be directly generated in a single step:
\begin{align}
    \rv{z}_t =  \alpha_t \rv{z}_0 + \gamma_t \rv{\epsilon},
\end{align}
where $\alpha_t := \prod_{i=1}^t \sqrt{1-\beta_i}$, $\gamma_t := \sqrt{1-\alpha_t^2}$, and $\rv{\epsilon} \sim \normaldist{0}{I}$.
The objective is to train a parametric model $\rv{\epsilon}_\rv{\zeta}(\rv{z}_t, t)$ by minimizing the score matching objective,

\begin{align}
    \label{eq:final-loss-diffusion}
    \mathcal{L}_\text{DDPM}(\rv{\zeta}) := \mathbb{E}_{p(\rv{z}_0),t \sim \mathcal{U}(1,T), \rv{\epsilon} \sim \normaldist{0}{I}} \left[ \| \rv{\epsilon} - \rv{\epsilon}_\rv{\zeta}(\rv{z}_t,t) \|_2^2 \right],
\end{align}
where $\mathcal{U}(1,T)$ is the uniform distribution on $\{1,2,\ldots,T\}$.

During inference, the network allows sampling through an iterative procedure since the learned distribution can be factorized as
\begin{equation}
\label{eq:ddpm-gen}
    p_\rv{\zeta}(\rv{z}) = p(\rv{z}_T) \ldm  = p(\rv{z}_T) \prod_{t=1}^T p_\rv{\zeta} (\rv{z}_{t-1} \vert \rv{z}_t)
\end{equation}
for $p(\rv{z}_T) := \normaldist{0}{I}$.

For a more detailed derivation, please refer to \cref{sec:supp-ddpm} in the supplementary.

\subsection{Spherical Harmonics}
\label{sec:background-Spherical harmonics}
The spherical harmonics~\cite{courant2008methods} are a set of base functions $Y_{l,m}: \unitsphere \rightarrow \mathbb{C}$ defined on a unit sphere $\unitsphere$ indexed by the degree $l \geq 0$ and the order $m$ ($-l \leq m \leq l)$.
They form an orthonormal basis for square-intergrable functions defined on the unit sphere $L^2(\unitsphere)$~\cite{rahaman2019spectral}; \ie, every such function $f$ can be represented as
\begin{equation}
\label{eq:spherical-harmonics-representation}
    f(\theta, \varphi) = \sum_{l = 0}^\infty \sum_{m = - l}^l c_{l,m} Y_{l,m}(\theta, \varphi).
\end{equation}
where the coefficients $c_{l,m}$ can be calculated by
\begin{equation}
\label{eq:spherical-harmonics-coefficients}
    c_{l,m} = \int_0^{2\pi} \int_0^\pi f(\theta, \varphi)\overline{Y_{l,m}(\theta, \varphi)} \sin(\theta) \mathrm{d}\theta \mathrm{d}\varphi.
\end{equation}

Similar to Fourier analysis~\cite{fourier1808memoire}, spherical harmonics cast spatial data into the spectral domain, allowing the extraction of frequency-specific information: coefficients $c_{l,m}$ with a higher degree measure the intensity of the original data in a higher-frequency domain. Please refer to \cref{sec:supp-spherical-harmonics} in the supplementary for more details about spherical harmonics.
\section{Formulation}
\label{sec:formulation}

\begin{figure*}
    \centering
    \begin{subfigure}[c]{0.64\textwidth}
    \includegraphics[width=\textwidth]{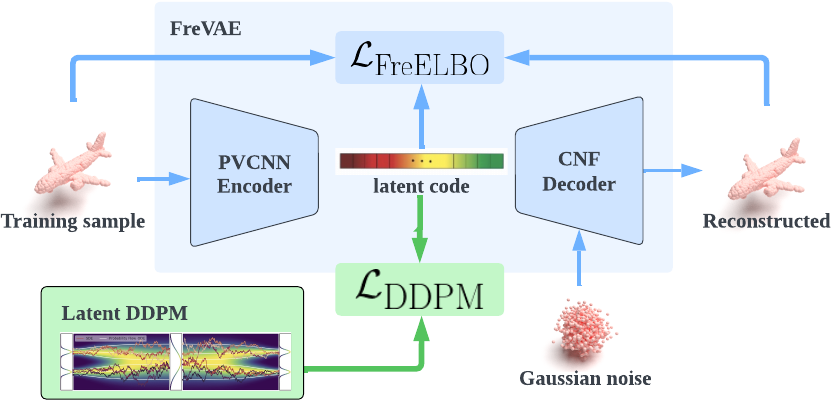}
    \caption{Two-stage training}
    \label{fig:arch-training}
    \end{subfigure}
    \hfill
    \begin{subfigure}[c]{0.35\textwidth}
    \includegraphics[width=\textwidth]{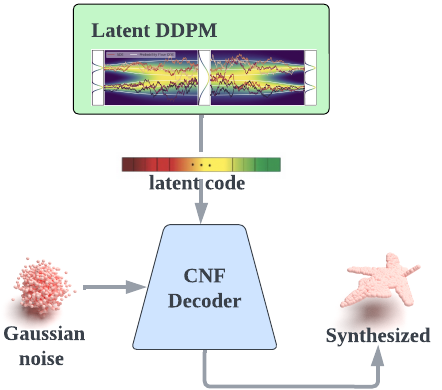}
    \caption{Generation}
    \label{fig:arch-gen}
    \end{subfigure}

    \caption{FrePolad is architectured as a point cloud VAE, with an embedded latent DDPM to represent the latent distribution. (a) Two-stage training: in the first stage (blue), the VAE is optimized to maximize the FreELBO \cref{eq:freelbo} with a standard Gaussian prior; in the second stage (green), while fixing the VAE, the latent DDPM is trained to model the latent distribution. (b) Generation: conditioned on a shape latent sampled from the DDPM, the CNF decoder transforms a Gaussian noise input into a synthesized shape.}
    \label{fig:architure}
\end{figure*}

We aim to learn a generative model to synthesize point clouds 
$\rv{X} \in \mathbb{R}^{3 \times N_\rv{X}}$
consisting of $N_\rv{X}$ points with a distribution $p(\rv{X})$.
Following~\cite{yang2019pointflow}, we treat each $\rv{X}$ as a distribution $p(\rv{x} \vert \rv{z})$ over its points $\rv{x} \in \mathbb{R}^3$, conditioned on the latent vector $\rv{z} \sim p(\rv{z})$ of the underlying shape.
Note that once $p(\rv{x} \vert \rv{z})$ is properly learned, the cardinality of the point cloud is unrestricted as one can sample an arbitrary number of points from this distribution.

\Cref{fig:architure} presents an overview of our generative model, \emph{\textbf{f}requency-\textbf{re}ctified \textbf{po}int \textbf{la}tent \textbf{d}iffusion (FrePolad)}. FrePolad amalgamates a point cloud VAE that models $p(\rv{x} \vert \rv{z})$ with a DDPM modeling its latent distribution $p(\rv{z})$. Synthesis is performed
by sampling a shape latent from the DDPM which is subsequently decoded into a point cloud. To train
the VAE, we leverage spherical harmonics and extract frequency information from point clouds (\cref{sec:formulation-Frequency Extraction via Spherical Harmonics}). Then, we formulate a novel frequency rectification module that encourages the reconstruction of high-frequency regions (\cref{sec:formulation-Frequency-Rectified VAE}). In addition, we employ a latent DDPM for enhanced expressiveness, continuous preservation of high-frequency contents, and reduced dimensionality. Collectively, we harmonize these components for our model, FrePolad, to facilitate point cloud generation (\cref{sec:formulation-frepolad}).

\subsection{Frequency Extraction via Spherical Harmonics}
\label{sec:formulation-Frequency Extraction via Spherical Harmonics}
In order to extract frequency information with spherical harmonics (\cref{eq:spherical-harmonics-representation}), we seek to represent the underlying surface of a point cloud $\rv{X} \in \mathbb{R}^{3 \times N_\rv{X}}$ consisting of $N_\rv{X}$ points by a continuous function on the unit sphere, \ie, $f_\rv{X}(\theta, \varphi)$.
Note that, with this representation, we assume the surface is a \emph{star domain}.
We start by expressing each 3D point in spherical coordinates $\rv{x}_i = (r_i, \theta_i, \varphi_i)$. Next, we construct $f_\rv{X}$ by anchoring each point $\rv{x}_i \in \rv{X}$ to its radius $r_i$ and interpolating in the rest of the domain:
\begin{align}
\label{eq:def-rep}
    f_\rv{X}(\theta, \varphi) := \begin{cases} 
      r & (\theta, \varphi, r) \in \rv{X} \\
      \sum\limits_{\begin{subarray}{c}
    (\theta_i, \varphi_i) \\
    \in \mathcal{H}_\rv{X}(\theta, \varphi)
\end{subarray}} w_{i}(\theta, \varphi) r_i & \textrm{otherwise}
   \end{cases}\,,
\end{align}
where $\mathcal{H}_\rv{X}(\theta, \varphi)$ is the set of $k$ closest neighbors $(\theta_i, \varphi_i)$ of $(\theta, \varphi)$ with $(r_i, \theta_i, \varphi_i) \in \rv{X}$ for some $r_i$, determined according to the following distance function $d$,
\begin{equation}
\begin{aligned}
\label{eq:repr-func}
    d_\text{sphere}((\theta, \varphi), (\theta', \varphi')) := \sqrt{2 - 2 \left[ \sin \theta \sin \theta' \cos(\varphi-\varphi') + \cos \theta \cos \theta'  \right]},
\end{aligned}
\end{equation}
and $w_{i}(\theta, \varphi)$ are the weights.

Since closer points should have larger weights, $\{w_i(\theta, \varphi)\}_i$ should be a decreasing sequence on $d_\text{sphere}((\theta, \varphi), (\theta_i, \varphi_i))$. A suitable candidate is the normalized Gaussian function with standard deviation $\sigma_\text{KNN}$:
\begin{align}
\label{eq:sh-knn-coeffs}
    & w'_{i}(\theta, \varphi) := e^{-\frac{d_\text{sphere}^2((\theta, \varphi), (\theta_i, \varphi_i))}{2 \sigma_\textrm{KNN}^2}} \, \textrm{and}\, w_{ij} := \frac{w_{ij}'}{\sum_j w_{ij}'}.
\end{align}

Defining $f_\rv{X}$ in this way presents limitations for point clouds with non-star-domain shapes. Specifically, at some values of $\theta$ and $\phi$, there may exist multiple points from semantically different regions of the point cloud. Nevertheless, we found this formulation to be computationally efficient and empirically superior to more complex alternatives (see the ablation study in \cref{sec:ablation-study} for more details).

After representing $\rv{X}$ by $f_\rv{X}$, we extract frequency information by computing its spherical harmonic coefficients $c_{l,m}^\rv{X}$ via \cref{eq:spherical-harmonics-coefficients}; these coefficients reveal the original point cloud $\rv{X}$ in frequency domains. We show a point cloud and its representative function in spherical and frequency domains in \cref{fig:pc-after-sh} (first row).

\subsection{Frequency-Rectified VAE}
\label{sec:formulation-Frequency-Rectified VAE}

As discussed earlier, VAEs suffer from losing high-frequency data, and existing remedies do not generalize well to point clouds~\cite{dosovitskiy2016generating, hou2017deep, esser2021taming, johnson2016perceptual, lee2022autoregressive}. To mitigate this problem, we utilize the frequency information extracted in \cref{sec:formulation-Frequency Extraction via Spherical Harmonics} and propose \emph{frequency rectification} in training the VAE that promotes the preservation of high-frequency information.


We first define a frequency-rectified distance $d_\textrm{Fre}(\rv{X}, \rv{X}')$ between two point clouds $\rv{X}$ and $\rv{X}'$:
\begin{equation}
\label{eq:new-loss-infinit-degree-xx}
    d_\text{Fre} \left(\rv{X}, \rv{X}' \right) := \sum_{l=0}^{\infty} \sum_{m=-l}^l r_l \left \|c_{l,m}^{\rv{X}} - c_{l,m}^{\rv{X}'} \right \|_2^2,
\end{equation}
where $\{ r_l \}_l$ is a sequence of increasing \emph{frequency rectifiers} that weight higher-degree spherical harmonic coefficients more. In practice, we restrict the infinite sum and evaluate the first $L + 1$ terms. The frequency rectifiers are given by the Gaussian function:
\begin{align}
\label{eq:freq-rectifiers}
    & r_{l} := e^{-\frac{(L - l)^2}{2 \sigma_\textrm{Fre}^2}}.
\end{align}

A frequency rectified loss between encoder and decoder distributions can be obtained by taking the expectation over point clouds $\rv{X}'$ learned by the VAE:
\begin{align}
\label{eq:new-loss-infinit-degree-distribution}
    \mathcal{L}_\text{Fre} \left(\rv{\psi}, \rv{\xi}; \rv{X} \right) := \mathbb{E}_{\rv{z} \sim \vaeenc, \rv{X}' \sim \vaedec} \left[ d_\text{Fre} \left( \rv{X}, \rv{X}' \right) \right]
\end{align}

In order to encourage the reconstruction of high-frequency regions, we introduce a constraint while maximizing the ELBO (\cref{eq:vae-elbo-no-beta}):
\begin{equation}
\begin{aligned}
\label{eq:abvae-optimization-problem}
    \underset{\rv{\psi}, \rv{\xi}}{\text{max}}\; & \mathbb{E}_{\rv{X} \sim p(\rv{X})} \left[ \mathcal{L}_\text{ELBO}(\rv{\psi}, \rv{\xi}; \rv{X}) \right] \\ \text{s.t.}\; & \mathbb{E}_{\rv{X} \sim p(\rv{X})} \left[ \mathcal{L}_\text{Fre} \left(\rv{\psi}, \rv{\xi}; \rv{X} \right) \right] < \delta,
\end{aligned}
\end{equation}
where $\delta > 0$ controls the strength of this constraint. Leveraging Karush–Kuhn–Tucker (KKT) methods~\cite{kuhn1951nonlinear, karush1939minima}, we can re-write the above constrained optimization problem (\ref{eq:abvae-optimization-problem}) as a Lagrange function whose optimal point is a global maximum over the domain of $( \rv{\psi}, \rv{\xi})$:
\begin{align}
    \mathcal{F}(\rv{\psi}, \rv{\xi}, \eta; \rv{X}) := \mathcal{L}_\text{ELBO}(\rv{\psi}, \rv{\xi}; \rv{X}) - \eta \left( \mathcal{L}_\text{Fre} \left(\rv{\psi}, \rv{\xi}; \rv{X}\right) - \delta \right),
\end{align}
where $\eta$ is the KKT multiplier. Since $\delta > 0$, we obtain a lower-bound on the ELBO:
\begin{equation}
\label{eq:freelbo}
\begin{aligned}
     \mathcal{F}(\rv{\psi}, \rv{\xi}, \eta; \rv{X})
    &\geq \mathcal{L}_\text{ELBO}(\rv{\psi}, \rv{\xi}; \rv{X}) - \eta \mathcal{L}_\text{Fre} \left(\rv{\psi}, \rv{\xi}; \rv{X}\right) \\
    & =: \mathcal{L}_\text{FreELBO}\left(\rv{\psi}, \rv{\xi}, \eta; \rv{X} \right).
\end{aligned}
\end{equation}

Our encoder and decoder networks are trained by maximizing the novel \emph{frequency-rectified evidence lowerbound (FreELBO)} $\mathcal{L}_\text{FreELBO}$. The hyperparameter $\eta$ trades-off reconstruction quality between the spatial and spectral domains. When $\eta=0$, the training objective is the same as the original ELBO \cref{eq:vae-elbo-no-beta}; when $\eta>0$, inaccuracies in reconstructing high-frequency regions are penalized.

\begin{figure}[tbp]
\centering
\includegraphics[width=0.6\linewidth]{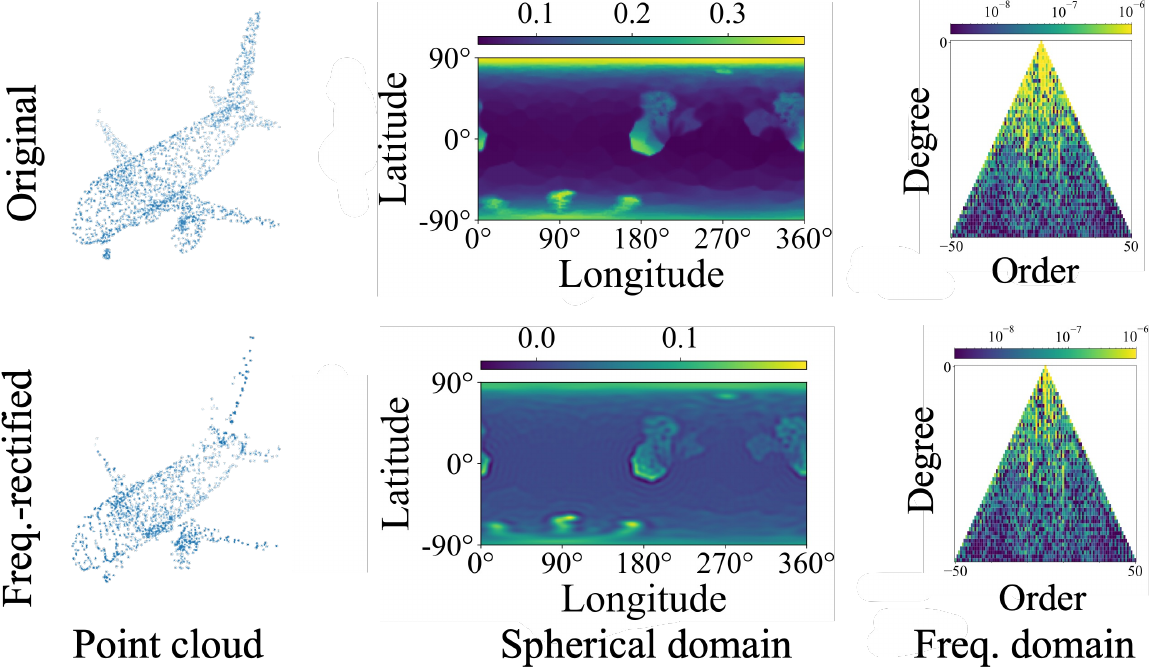}
\caption{A point cloud before and after frequency rectification and its representative function in spherical and frequency domains. Frequency rectification shifts points to more complex, less smooth regions and increases the relative importance of higher-frequency features, where VAEs can give more attention during reconstruction. Note that the frequency rectified point cloud in the second row is only for visualization; our framework does not explicitly generate such a point cloud during training.}
\label{fig:pc-after-sh}
\end{figure}

An example of a frequency-rectified point cloud is shown in \cref{fig:pc-after-sh}. We observe that after frequency rectification, points shift to more complex or less smooth regions in the point clouds, corresponding to higher frequency areas. Additionally, in the frequency domain, we see that as lower-frequency features (with lower degrees) are decayed by the rectifiers \cref{eq:freq-rectifiers}, the relative significance of higher-frequency features amplifies. Consequently, our VAE prioritizes the reconstruction in these regions.

\subsection{DDPM-Based Prior}
\label{sec:formulation-Point latent diffusion}
Although it is possible to sample shape latents from a simple Gaussian prior for generation, evidence suggests that such a restricted prior cannot accurately capture complex encoder distributions, $\vaeenc$. This is known as the \emph{prior hole problem}~\cite{vahdat2021score, tomczak2018vae, rosca2018distribution}, and tends to curtail the performance of VAEs~\cite{chen2016variational}, resulting in the poor reconstructions in our ablation study (see \cref{sec:ablation-study}).
To address this, we employ a DDPM to model the VAE's latent distribution~\cite{rombach2022high}.
The DDPM is trained on the latents $\rv{z}$ directly sampled from $\vaeenc$.
Since the original prior is insufficient, the DDPM learns a more expressive distribution $p_\zeta(\rv{z})$, better matches the true latent distribution $p(\rv{z})$, and functions as the VAE's prior.
Combining latent DDPM with a VAE-based model helps achieve a near-ideal equilibrium between minimizing complexity and ensuring perceptual fidelity~\cite{vahdat2021score, rombach2022high}. Crucially, the high-frequency details maintained by frequency rectification during encoding stage remain undisturbed. Additionally, learning the distribution of shape latents instead of point clouds significantly reduces the training and sampling cost (see \cref{tab:exp-time}) due to the reduced dimensionality. This enables scalable generation of much denser point clouds (see \cref{fig:flex-gen}).

\subsection{FrePolad}
\label{sec:formulation-frepolad}

With all the building blocks, we present our new generative model for point cloud generation: \emph{FrePolad: \textbf{f}requency-\textbf{re}ctified \textbf{po}int \textbf{la}tent \textbf{d}iffusion}. The overall structure of our framework is presented in \cref{fig:architure}.

\paragraph{Components}
The VAE encoder $\vaeenc$ is a Point-Voxel CNN (PVCNN)~\cite{liu2019point, zhou20213d} parametrized by $\rv{\psi}$. PVCNNs efficiently combine the point-based processing of PointNets~\cite{qi2017pointnet, qi2017pointnet++} with the strong spatial inductive bias of convolutions. Our encoder accommodates point clouds of variable cardinalities and is permutation-invariant.

In order to support flexible cardinality while synthesizing point clouds, we interpret each point cloud $\rv{X}$ as a distribution $p(\rv{x} \vert \rv{z})$ of its constituent points conditioned on the shape latent $\rv{z}$. Assuming independence among the points in a point cloud, we model the decoder distribution as
\begin{equation}
\label{eq:decoder-model}
\vaedec := \prod_{\rv{x} \in \rv{X}} p_\rv{\xi}(\rv{x} \vert \rv{z}).
\end{equation}

The decoder is implemented using a conditional continuous normalizing flow~(CNF)~\cite{rezende2015variational, grathwohl2018ffjord, chen2018neural}. Here, a sampled point $\rv{x}$ is the result of transforming some initial point $\rv{x}(0) \sim p(\rv{x}(0)) := \normaldist{0}{I}$, conditioned on the shape latent $\rv{z}$. The invertible nature of CNFs offers a high degree of interpretability and also allows the exact computation of data likelihood $p(\rv{x} \vert \rv{z})$ by moving the transformed points back to the prior distribution. For more details, please see \cref{sec:supp-cnf} in the supplementary.

The latent DDPM $p_\rv{\zeta}(\rv{z})$ is parametrized by $\rv{\zeta}$, realized through a U-Net backbone~\cite{ronneberger2015u} following~\cite{rombach2022high}.

\paragraph{Training}
Following common practice~\cite{rombach2022high, esser2021taming, razavi2019generating, cai2020learning}, we perform a two-stage training. In the first stage, we train the VAE network by maximizing the FreELBO \cref{eq:freelbo} with the prior $p(\rv{z}) := \normaldist{0}{I}$. In the second stage, we freeze the VAE network and train the DDPM on the latent vectors sampled from the encoder $\vaeenc$ by minimizing the objective function \cref{eq:final-loss-diffusion}.

\paragraph{Generation}
During generation, new point clouds of arbitrary cardinality can be generated by first sampling a shape latent $\rv{z}$ from the DDPM $p_\rv{\zeta}(\rv{z})$ following \cref{eq:ddpm-gen} and then sampling the decoder $\vaedec$ following \cref{eq:decoder-model}:
Formally, the generation process of FrePolad is defined by
\begin{align}
    p_\rv{\xi,\zeta}(\rv{X}) := p_\rv{\zeta}(\rv{z}) \vaedec.
\end{align}


\section{Experiments}
\label{sec:experiments}
In this section, we provide experiments demonstrating FrePolad's performance in point cloud generation. Please refer to \cref{sec:supp-hyperparameters} in the supplementary for training and hyperparameter details.

\begin{figure*}
\centering
\includegraphics[width=\linewidth]{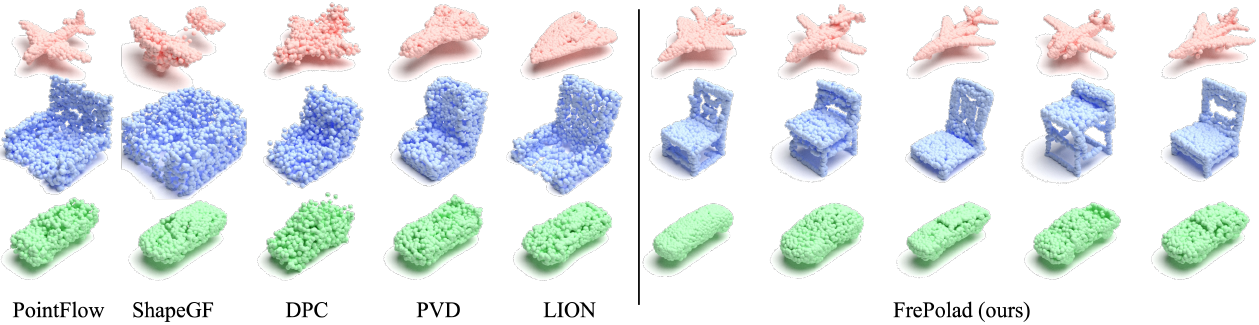}
\caption{Generation with 2048 points for airplane, chair, and car classes. Samples generated by FrePolad have better fidelity and diversity.}
\label{fig:exp-gen}
\end{figure*}

\subsection{Dataset}
To benchmark FrePolad against existing methods, we employ ShapeNet~\cite{chang2015shapenet}, a widely used dataset for 3D generative task assessments.
Following prior work~\cite{vahdat2022lion, luo2021diffusion, yang2019pointflow, zhou20213d}, we train on three categories: \emph{airplane}, \emph{chair}, and \emph{car}. We follow the dataset splits and preprocessing established in PointFlow~\cite{yang2019pointflow}. Unless stated otherwise, each point cloud consists of 2048 points for each shape.

\subsection{Evaluation metrics}
In line with previous works~\cite{vahdat2022lion, luo2021diffusion, zhou20213d, yang2019pointflow}, we employ 1-nearest neighbor (1-NNA)~\cite{lopez2016revisiting}
, computed using both Chamfer distance (CD) and earth mover distance (EMD). 1-NNA assesses the similarity between two 3D shape distributions, considering both diversity and quality~\cite{yang2019pointflow}. \Cref{sec:supp-full-table} in the supplementary provides further comparisons with two other metrics: minimum matching distance (MMD)~\cite{achlioptas2018learning} and coverage (COV)~\cite{achlioptas2018learning}.

\subsection{Results}
\begin{figure}
\centering
\begin{minipage}{0.48\textwidth}
\includegraphics[width=\linewidth]{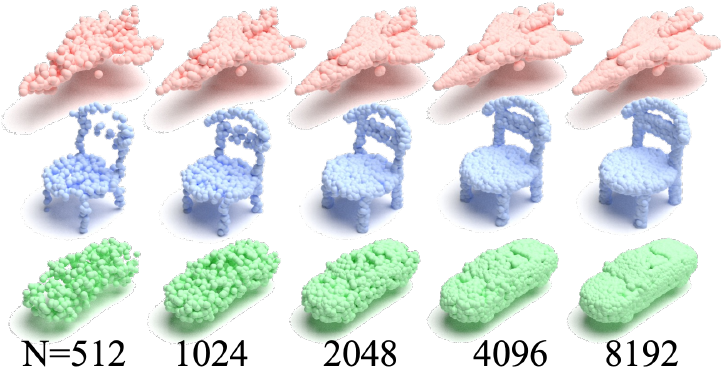}
\caption{FrePolad supports flexibility in the cardinality of generated point clouds.}
\label{fig:flex-gen}
\end{minipage}
\hfill
\begin{minipage}{0.48\textwidth}
\hfill
\includegraphics[width=\linewidth]{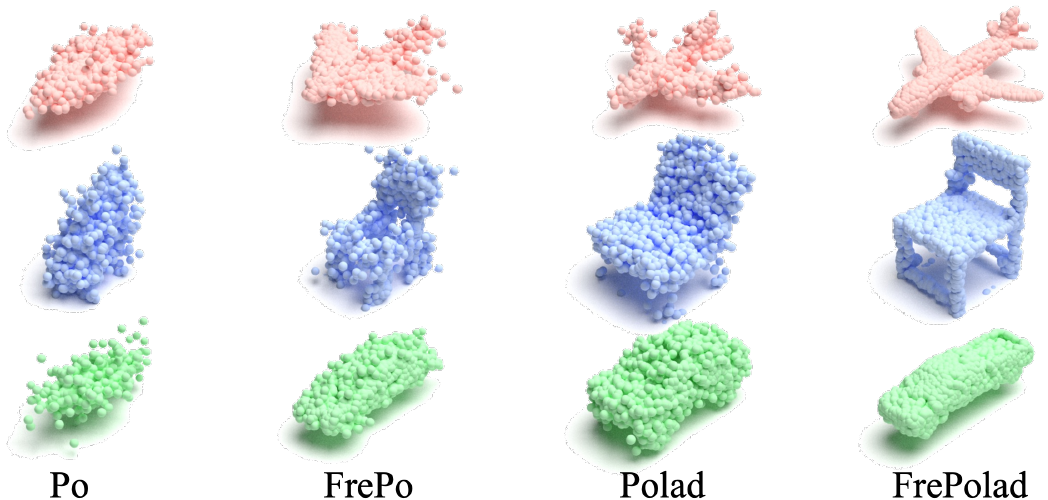}
\caption{Ablation study: shape generation by three simplified versions of FrePolad.}
\label{fig:exp-ablation}
\end{minipage}
\end{figure}

\paragraph{Point cloud generation}
In \cref{tab:exp-table}, we benchmark our model, FrePolad, against several recent works.
Demonstrating high generation fidelity and diversity, our model consistently surpasses most baselines and approaches the ground truth (training set).
Importantly, FrePolad attains excellent results comparable to the state-of-the-art model LION~\cite{vahdat2022lion} despite a simpler architecture that is computationally efficient during both training and inference (see the efficiency evaluation below). Further, by modeling each point cloud as a point distribution over a shape latent, our method facilitates sampling point clouds with arbitrary cardinality
-- a feature rarely available in other baselines. \Cref{fig:flex-gen} illustrates the generation with different numbers of points.

\Cref{fig:exp-gen} provides qualitative comparisons, highlighting the visually appealing and diverse point clouds generated by FrePolad across three distinct classes.

\begin{table}
\centering
{
\small
\setlength\tabcolsep{3.5pt}
\begin{tabular}{lcccccc}
\toprule
\multicolumn{1}{c}{\multirow{2}{*}{Model}} & \multicolumn{2}{c}{Airplane} & \multicolumn{2}{c}{Chair} & \multicolumn{2}{c}{Car} \\
\cmidrule(lr){2-3} \cmidrule(lr){4-5} \cmidrule(lr){6-7}
\multicolumn{1}{c}{} & \multicolumn{1}{c}{CD} & \multicolumn{1}{c}{EMD} & \multicolumn{1}{c}{CD} & \multicolumn{1}{c}{EMD} & \multicolumn{1}{c}{CD} & \multicolumn{1}{c}{EMD} \\
\midrule
Training set & 64.44 & 64.07 & 51.28 & 54.76 & 51.70 & 50.00 \\
\cmidrule{2-7}
 r-GAN~\cite{achlioptas2018learning} & 98.40 & 96.79 & 83.69 & 99.70 & 94.46 & 99.01 \\
 1-GAN/CD~\cite{achlioptas2018learning} & 87.30 & 93.95 & 68.58 & 83.84  & 66.49 & 88.78 \\
 1-GAN/EMD~\cite{achlioptas2018learning} & 89.49 & 76.91 & 71.90 & 64.65 & 71.16 & 66.19 \\
 PointFlow~\cite{yang2019pointflow} & 75.68 & 70.74 & 62.84 & 60.57 & 58.10 & 56.25 \\
 SoftFlow~\cite{kim2020softflow} & 76.05 & 65.80 & 59.21 & 60.05 & 64.77 & 60.09 \\
 SetVAE~\cite{kim2021setvae} & 76.54 & 67.65 & 58.84 & 60.57 & 59.94 & 59.94 \\
 ShapeGF~\cite{cai2020learning} & 81.23 & 80.86 & 58.01 & 61.25 & 61.79 & 57.24 \\
 DPF-Net~\cite{klokov2020discrete} & 75.18 & 65.55 & 62.00 & 58.53 & 62.35 & 54.48 \\
 DPC~\cite{luo2021diffusion} & 76.42 & 86.91 & 60.05 & 74.77 & 68.89 & 79.97 \\
 PVD~\cite{zhou20213d} & 73.82 & 64.81 & 56.26 & 53.32 & 54.55 & 53.83 \\
 LION~\cite{vahdat2022lion} & 67.41 & \textbf{61.23} & 53.70 & \textbf{52.34} & 53.41 & 51.14 \\
 FrePolad (ours) & \textbf{65.25} & 62.10 & \textbf{52.35} & 53.23 & \textbf{51.89} & \textbf{50.26} \\
 \bottomrule
\end{tabular}
}
\caption{1-NNA scores ($\downarrow$) on point cloud generation on three classes of ShapeNet dataset. FrePolad achieves significant improvement over most existing models and obtains state-of-the-art results with simpler and more flexible structure.}
\label{tab:exp-table}
\end{table}

\begin{figure}
    \centering
    \includegraphics[width=0.6\linewidth]{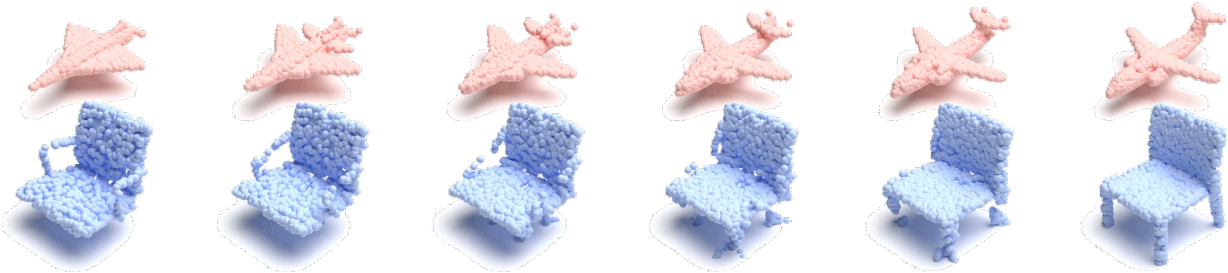}
    \caption{Interpolation of shapes in the VAE latent space.}
    \label{fig:interpolation}
\end{figure}

\paragraph{Latent interpolation}  \cref{fig:interpolation} showcases latent interpolations facilitated by FrePolad. The smooth transitions in the interpolated shapes suggest that our network successfully captures an effective latent space of the underlying shapes.

\paragraph{Ablation study}
\label{sec:ablation-study}
In \cref{tab:exp-ablation} and \cref{fig:exp-ablation} we also include quantitative and qualitative comparisons of three simplified variants of FrePolad. We selectively omit some components such as the frequency rectification and/or the latent DDPM:
\begin{itemize}
    \item Polad (FrePolad without \textbf{f}requency \textbf{re}ctification): in the first stage of training, the VAE is trained with the standard ELBO \cref{eq:vae-elbo-no-beta} in lieu of the FreELBO \cref{eq:freelbo};

    \item FrePo (FrePolad without \textbf{la}tent \textbf{d}iffusion): instead of using a latent diffusion as the latent distribution of VAE, we assume the prior is a standard Gaussian; and

    \item Po (FrePolad without both).
\end{itemize}
Our results indicate that incorporating frequency rectification and latent DDPM substantially enhances the quality and diversity of the generated point clouds.

\begin{table}
\centering
{
\small
\setlength\tabcolsep{5pt}
\begin{tabular}{lcccccc}
\toprule
\multicolumn{1}{c}{\multirow{2}{*}{Model}} & \multicolumn{2}{c}{Airplane} & \multicolumn{2}{c}{Chair} & \multicolumn{2}{c}{Car} \\
\cmidrule(lr){2-3} \cmidrule(lr){4-5} \cmidrule(lr){6-7}
\multicolumn{1}{c}{} & \multicolumn{1}{c}{CD} & \multicolumn{1}{c}{EMD} & \multicolumn{1}{c}{CD} & \multicolumn{1}{c}{EMD} & \multicolumn{1}{c}{CD} & \multicolumn{1}{c}{EMD} \\
\midrule
 Po & 85.93 & 85.94 & 94.53 & 99.22 & 100.0 & 100.0 \\
 FrePo & 80.38 & 80.38 & 89.06 & 82.19 & 89.22 & 84.53 \\
 Polad & 69.26 & 65.26 & 57.35 & 61.52 & 58.12 & 56.13 \\
 FrePolad & \textbf{65.25} & \textbf{62.10} & \textbf{52.35} & \textbf{53.23} & \textbf{51.89} & \textbf{50.26} \\
 \bottomrule
\end{tabular}
}
\caption{Ablation study: 1-NNA scores ($\downarrow$) of three simplified versions of FrePolad. The frequency rectification and the latent DDPM bring significant improvement.}
\label{tab:exp-ablation}
\end{table}

\paragraph{Computational efficiency}
\label{sec:exp-computational-efficiency}
We compare the computational efficiency of different models trained on ShapeNet airplane dataset~\cite{chang2015shapenet}. In \cref{tab:exp-time} and \cref{fig:teaser}~(b) to (e), we document the computational overhead incurred by different methods during training and generation.
From these results, it is evident that FrePolad exhibits a favorable computational efficiency. In contrast, considering other latent-based models, PointFlow \cite{yang2019pointflow} is much slower during training, likely due to its two flows (for the prior and decoder) trained jointly, whereas FrePolad only has a single flow for the decoder and we separate the training into two stages. LION \cite{vahdat2022lion} has a hierarchical latent space with a dimensionality larger than that of the original point cloud, leading to excessive computational cost during both training and sampling. During generation, FrePolad also achieves scalability while maintaining high  quality. FrePolad’s generation time remains constant as the bulk of time is spent on the latent diffusion model, unaffected by varying output point numbers due to the consistent latent dimension. The difference becomes even more evident when generating dense point clouds containing as many as 100k points.

\begin{table}
\centering
{
\small
\begin{tabular}{lccccc}
\toprule
\multicolumn{1}{c}{\multirow{2}{*}{Model}} & \multicolumn{1}{c}{Train} & \multicolumn{4}{c}{Generation (seconds per shape)} \\
\cmidrule{3-6}
\multicolumn{1}{c}{} & \multicolumn{1}{c}{(hours)} & \multicolumn{1}{c}{2048} & \multicolumn{1}{c}{8192} & \multicolumn{1}{c}{15k} & \multicolumn{1}{c}{100k} \\
\midrule
PointFlow~\cite{yang2019pointflow} & 360 & 1.73 & 1.76 & 1.77 & 1.81 \\
ShapeGF~\cite{cai2020learning} & 28 & 20.60 & 20.71 & 20.87 & 21.42 \\
DPC~\cite{luo2021diffusion} & 300 & 27.34 & 110 & 220 & 1370 \\
PVD~\cite{zhou20213d} & 230 & 200 & - & - & - \\
LION~\cite{vahdat2022lion} & 550 & 30.2 & - & - & - \\
FrePolad (ours) & \textbf{20} & \textbf{6.30} & \textbf{6.32} & \textbf{6.35} & \textbf{6.41} \\
 \bottomrule
\end{tabular}
}
\caption{Training cost on the airplane dataset and generation with different point cloud cardinalities. ``-'' means unsupported. FrePolad incurs most efficient computational cost and can be easily scaled up to generate significantly denser point clouds.}
\label{tab:exp-time}
\end{table}





\subsection{Direct generation \textbf{\vs} upsampling}
\label{sec:upsampling}
The flexibility of FrePolad enables the generation of arbitrarily dense point clouds. While such dense point clouds can also be acquired by upsampling sparse ones, we claim that FrePolad's direct generation approach yields superior quality. To demonstrate this, we consider the generation of point clouds of different cardinalities containing 8192, 15k, and 100k points. We consider a few baselines that generate dense point clouds by upsampling sparser ones using recent, competitive methods.
Due to the GPU memory constraints, we only sample 8192 points when evaluating 1-NNA-CD. The result is summarized in \cref{tab:exp-upsampling-compare} (further results with other data classes or metrics can be found in \cref{tab:exp-upsampling-compare-big} in the supplementary). We can observe that the direct generation by FrePolad vastly outperforms the upsampling approach in terms of generation quality and runtime. This comparison clearly highlights the advantage of FrePolad's capability to directly generate dense point clouds.
\begin{table}
\centering
{
\small
\setlength\tabcolsep{5pt}
\begin{tabular}{lcccccc}
\toprule
\multicolumn{1}{c}{\multirow{2}{*}{Model}} & \multicolumn{3}{c}{1-NNA-CD ($\downarrow$)} & \multicolumn{3}{c}{Time (min)} \\
\cmidrule(lr){2-4} \cmidrule(lr){5-7}
\multicolumn{1}{c}{} & \multicolumn{1}{c}{8192} & \multicolumn{1}{c}{15k} & \multicolumn{1}{c}{100k} & \multicolumn{1}{c}{8192} & \multicolumn{1}{c}{15k} & \multicolumn{1}{c}{100k} \\
\midrule
 PU-GAN~\cite{li2019pu} & 100 & 100 & 100 & 1.17 & 2.44 & 7.55 \\
 PU-GCN~\cite{qian2021pu} & 100 & 100 & 100 & 0.83 & 6.67 & 96 \\
 Grad-PU~\cite{he2023grad} & 100 & 100 & 100 & 0.87 & 2.47 & 68.52 \\
 \midrule
 FrePolad (ours) & \textbf{65.31} & \textbf{66.87} & \textbf{66.25} & \textbf{0.11} & \textbf{0.11} & \textbf{0.11} \\
 \bottomrule
\end{tabular}
}
\caption{1-NNA-CD ($\downarrow$) and time cost of the generation of dense point clouds with different cardinalities. The first three methods upsample sparse ground truths while FrePolad directly generates dense ones. FrePolad substantially outperforms all upsampling methods in terms of generation quality and runtime, demonstrating the importance of the ability to directly generate dense point clouds.}
\label{tab:exp-upsampling-compare}
\end{table}
\section{Conclusion and Future Works}
In this paper, we presented FrePolad, a novel method for point cloud generation. At its core, FrePolad is structured as a point cloud VAE integrated with a DDPM-based prior.
To enhance sample quality and diversity, we tailor a new frequency rectification technique that preserves the high-frequency details extracted via spherical harmonics. We also incorporate a latent DDPM to model the regularized latent space of the VAE. Additionally, to achieve flexibility in the point cloud cardinality, we perceive point clouds as distributions of their constituent points over a latent shape.
Our proposed model exhibits minimal training and sampling overheads and can be easily scaled to generate highly dense point clouds. Our empirical analyses, both quantitative and qualitative, showed the state-of-the-art performance achieved by FrePolad.

We currently use the standard spherical harmonic base functions $Y_{l,m}$ in \cref{eq:spherical harmonic base functions} on the unit sphere for frequency analysis.
An interesting future direction is to investigate a more general differential analysis on manifolds where the base functions are defined on the smoothed surface derived from the original point cloud. Additionally, we aspire to explore the application of frequency rectification in other domains with complex signals containing high-frequency information.

\setcounter{secnumdepth}{0}
\section{Acknowledgement}
This work was supported by a UKRI Future Leaders Fellowship [grant number G104084].

%
%
\bibliographystyle{splncs04}
\bibliography{main}

\clearpage
\setcounter{page}{1}
\setcounter{section}{0}
\renewcommand{\thesection}{\Alph{section}}
\maketitlesupplementary
\section{Denoising Diffusion Probabilistic Model}
\label{sec:supp-ddpm}
Given a data sample $\rv{z} \sim q(\rv{z})$, at each time step $t=1,2,\ldots,T$, DDPMs~\cite{ho2020denoising, sohl2015deep} gradually transform $\rv{z} =\rv{z}_0$ into $\rv{z}_T$ by adding noise in a Markovian \emph{diffusion process} defined by
\begin{align}
q \left( \rv{z}_{1:T} \vert \rv{z}_0 \right) &:= \prod_{t=1}^T q ( \rv{z}_t \vert \rv{z}_{t-1} );\\
\label{eq:q-zt-given-zt-1}
q(\rv{z}_t \vert \rv{z}_{t-1}) &:= \normaldist{\sqrt{1 -\beta_t} \rv{z}_{t-1}}{\beta_t I},
\end{align}
where $\normaldist{\rv{\mu}}{\rv{\sigma}}$ denotes multivariate Gaussian distribution with mean $\rv{\mu}$ and variance $\rv{\sigma}$, and $\{\beta_t\}_t$ is a pre-determined variance schedule controlling the rate of diffusion. If $T$ is sufficiently large (1000 steps in practice), $p(\rv{z}_T)$ approaches the standard Gaussian.

DDPMs learn an inverse Markovian process called \emph{reverse process} parametrized by $\rv{\zeta}$ that inverts the forward diffusion, transforming the standard Gaussian noise $\rv{z}_T$ back to a data sample:
\begin{align}
\label{eq:markov}
p_\rv{\zeta}(\rv{z}_{0:T}) &:= p(\rv{z}_T) \prod_{t=1}^T p_\rv{\zeta}(\rv{z}_{t-1} \vert \rv{z}_t);\\
\label{eq:p-zt-1-given-zt}
p_\rv{\zeta}(\rv{z}_{t-1} \vert \rv{z}_t) &:= \normaldist{\mu_\rv{\zeta}(\rv{z}_t, t)}{\sigma_t^2 I},
\end{align}
where $\mu_\rv{\zeta} (\rv{z}_t, t)$ represents the predicted mean for the Gaussian distribution at time step $t$ and $\{\sigma_t\}_t$ is the variance schedule.

DDPMs are trained by maximizing the variational lower bound of log-likelihood of the data $\rv{z}_0$ under $q(\rv{z}_0)$:
\begin{align}
    \label{eq:variational-lower-bound}
    \mathbb{E}_{q(\rv{z}_0)} \left[ \log p_\rv{\zeta}(\rv{z}_0) \right] \geq \mathbb{E}_{q(\rv{z}_{0:T})} \left[ \log \frac{p_\rv{\zeta}(\rv{z}_{0:T})}{q(\rv{z}_{1:T} \vert \rv{z}_0)} \right].
\end{align}
Expanding \cref{eq:variational-lower-bound} with \cref{eq:markov} and noticing that $p(\rv{z}_T)$ and $q(\rv{z}_{1:T} \vert \rv{z}_0)$ are constant with respect to $\rv{\zeta}$, we obtain our objective function to maximize:
\begin{align}
\label{eq:pre-objective}
    \mathbb{E}_{q(\rv{z}_0),q(\rv{z}_{1:T} \vert \rv{z}_0)} \left[ \sum_{t=1}^T \log p_\rv{\zeta} ( \rv{z}_{t-1} \vert \rv{z}_t) \right].
\end{align}
Since we can factor the joint posterior
\begin{equation}
q(\rv{z}_{1:T} | \rv{z}_0) = \prod_{t=1}^T q(\rv{z}_{t-1} \vert \rv{z}_t, \rv{z}_0)
\end{equation}
and both $q(\rv{z}_{t-1} \vert \rv{z}_t, \rv{z}_0)$ and $p_\rv{\zeta}(\rv{z}_{t-1} | \rv{z}_t)$ are Gaussian, maximizing \cref{eq:pre-objective} can be reduced to minimizing the following loss:
\begin{align}
    \label{eq:final-loss-diffusion-supp}
    \mathcal{L}_\text{DDPM}(\rv{\zeta}) := \mathbb{E}_{q(\rv{z}_0),t \sim \mathcal{U}(1,T), \rv{\epsilon} \sim \normaldist{0}{I}} \left[ \| \rv{\epsilon} - \rv{\epsilon}_\rv{\zeta}(\rv{z}_t,t) \|_2^2 \right],
\end{align}
where $\mathcal{U}(1,T)$ is the uniform distribution on $\{1,2,\ldots,T\}$. Note that here we employ a commonly used parametrization
\begin{align}
    \mu_\rv{\zeta} (\rv{z}_t, t) = \frac{1}{\sqrt{1-\beta_t}} \left( \rv{z}_t - \frac{\beta_t}{\gamma_t} \rv{\epsilon}_\rv{\zeta}(\rv{z}_t, t) \right)
\end{align}
and that via \cref{eq:q-zt-given-zt-1} $\rv{z}_t$ is in fact tractable from the initial $\rv{z}_0$ by
\begin{align}
    \rv{z}_t =  \alpha_t \rv{z}_0 + \gamma_t \rv{\epsilon},
\end{align}
where $\alpha_t := \prod_{i=1}^t \sqrt{1-\beta_i}$ and $\gamma_t := \sqrt{1-\alpha_t^2}$. Intuitively, minimizing the loss \cref{eq:final-loss-diffusion-supp} amounts to predict the noise $\rv{\epsilon}$ necessary to denoise the diffused sample $\rv{z}_t$.

During inference time, DDPMs can be iteratively sampled with ancestral sampling by first sampling $\rv{z}_T$ from $p(\rv{z}_T) := \normaldist{0}{I}$ and following \cref{eq:p-zt-1-given-zt}; that is:
\begin{equation}
\label{eq:ddpm-gen-supp}
    p_\rv{\zeta}(\rv{z}) = p(\rv{z}_T) \ldm  = p(\rv{z}_T) \prod_{t=1}^T p_\rv{\zeta} (\rv{z}_{t-1} \vert \rv{z}_t)
\end{equation}
for $p(\rv{z}_T) := \normaldist{0}{I}$.

\section{Spherical Harmonics}
\label{sec:supp-spherical-harmonics}
The spherical harmonics~\cite{courant2008methods} are a set of complex-valued spherical harmonic base functions $Y_{l,m}: \unitsphere \rightarrow \mathbb{C}$ defined on a unit sphere $\unitsphere$, indexed by the degree $l \geq 0$ and order $m$ ($-l \leq m \leq l)$:
\begin{equation}
\label{eq:spherical harmonic base functions}
    Y_{l,m}(\theta, \varphi) := Z_{l,m} e^{im\varphi} P_{l,m} \left(\cos \theta \right),
\end{equation}
where the colatitud $\theta \in [0, \pi]$, the longitude $\varphi \in [0, 2\pi)$, $Z_{l,m} \in \mathbb{C}$ is the normalization constant, $i$ is the imaginary unit, and $P_{l,m}: [-1, 1] \rightarrow \mathbb{R}$ is the associated Legendre polynomial of degree $l$ and order $m$ satisfying the general Legendre equation:

\small
\begin{equation}
\begin{aligned}
    &\frac{\dd}{\dd x} \left[ (1-x^2) \frac{\dd}{\dd x} P_{l,m}(x) \right]  + \left[ l(l+1) - \frac{m^2}{1-x^2} \right] P_{l,m}(x) = 0.
\end{aligned}
\end{equation}
\normalsize

Note that for a fixed degree $l$, any $Y_{l,m}$ for $-l \leq m \leq l$ are solutions to the differential equation
\begin{equation}
\label{eq:sh-laplace}
    \Delta Y_{l,m} = -l(l+1)Y_{l,m},
\end{equation}
where $\Delta$ is the Laplace operator, and that every solution to \cref{eq:sh-laplace} is a linear combination of $Y_{l,m}$, $-l \leq m \leq l$. In other words, spherical harmonics base functions are eigenfunctions of the Laplace operator (or, more generally in higher dimensions, the Laplace-Beltrami operator).

The spherical harmonics are a complete set of orthonormal functions and thus form an orthonormal basis for square-intergrable functions defined on the unit sphere $L^2(\unitsphere)$~\cite{rahaman2019spectral}; \ie, every continuous function $f: \unitsphere \rightarrow \mathbb{C}$ such that 
\begin{equation}
    \int_0^{2\pi} \int_0^\pi | f(\theta, \varphi) |^2 \mathrm{d}\theta \mathrm{d} \varphi < \infty
\end{equation}
can be represented as a series of these spherical harmonic base functions $Y_{l,m}$ by
\begin{equation}
\label{eq:supp-spherical-harmonics-representation}
    f(\theta, \varphi) = \sum_{l = 0}^\infty \sum_{m = - l}^l c_{l,m} Y_{l,m}(\theta, \varphi),
\end{equation}
where the coefficients $c_{l,m}$ can be calculated by
\begin{equation}
\label{eq:supp-spherical-harmonics-coefficients}
    c_{l,m} = \int_0^{2\pi} \int_0^\pi f(\theta, \varphi)\overline{Y_{l,m}(\theta, \varphi)} \sin(\theta) \mathrm{d}\theta \mathrm{d}\varphi.
\end{equation}

This is parallel to the result from Fourier analysis~\cite{fourier1808memoire} that any arbitrary function defined on a plane can be expressed as a trigonometric series. The trigonometric functions in a Fourier series capture the fundamental modes of vibration on a plane whereas the spherical harmonics denote these modes of vibration on a sphere. Consequently, spherical harmonics cast spatial data into the spectral domain, allowing the extraction of frequency-specific information: coefficients $c_{l,m}$ with a higher degree measure the intensity of the original data in a higher-frequency domain. Please refer to \cref{sec:supp-spherical-harmonics} in the supplementary for more details about spherical harmonics.

\section{Continuous Normalizing Flow}
\label{sec:supp-cnf}
Normalizing flows~\cite{rezende2015variational} consist of a sequence of reversible mappings $\{g^i\}_{i=1}^{n}$. It assumes that the data points $\rv{x} = \rv{x}_n$ are obtained by iteratively transforming a sample from an initial distribution $p(\rv{x}_0)$:
\begin{equation}
    \rv{x}_n := (g^n \circ g^{n-1} \circ \cdots \circ g^1)(\rv{x}_0).
\end{equation}
The probability density of the resultant variable $\rv{x}_n$ can be determined using the change of variables formula:
\begin{equation}
\label{eq:supp-nf-data-likelyhood}
    \log p(\rv{x}_n) = \log p(\rv{x}_0) - \sum_{i=1}^n \log \left\vert \det \left( \frac{\partial g^i(\rv{x}_{i-1})}{\partial \rv{x}_{i-1}} \right) \right\vert,
\end{equation}
where $\rv{x}_0$ can be computed from $\rv{x}_n$ using the inverse flow
\begin{equation}
    \rv{x}_0 = (g^1 \circ \cdots \circ g^{n-1} \circ g^n)(\rv{x}_n),
\end{equation}
and $\det(\cdot)$ is the Jacobian determinant function.

In practice, the initial distribution $p(\rv{x}_0)$ is chosen to be the standard Gaussian $\normaldist{0}{I}$, and the invertible mappings $\{g^i\}_{i=1}^{n}$ are represented by neural networks $\{g^i_\rv{\xi}\}_{i=1}^{n}$ parameterized by $\rv{\xi}$ for which the Jacobian determinant $\left\vert \det \left( \frac{\partial g_\rv{\xi}^i(\rv{x}_{i-1})}{\partial \rv{x}_{i-1}} \right) \right\vert$ is easy to compute. Since the exact log-likelihood of input data is tractable via \cref{eq:supp-nf-data-likelyhood}, the training of normalizing flows just involves maximizing \cref{eq:supp-nf-data-likelyhood}.

Normalizing flows can be generalized to continuous normalizing flows (CNFs) \cite{grathwohl2018ffjord, chen2018neural}, where a sequence of mappings $\{g_\rv{\xi}(\cdot, t)\}_{t=0}^\tau$ indexed by a real number $t \in \mathbb{R}$ transforms an initial point $\rv{x}(0) \sim p(\rv{x}_0)$ into $\rv{x}(\tau)$ following a continuous-time dynamic
\begin{equation}
    \frac{\partial \rv{x}_t}{\partial t} = g_\rv{\xi}(\rv{x}(t), t),
\end{equation}
and therefore
\begin{equation}
\label{eq:supp-cnf-dynamic}
    \rv{x}(\tau) = \rv{x}(0) + \int_0^\tau g_\rv{\xi}(\rv{x}(0), t) \dd t.
\end{equation}
Under this formulation, the probability density function $p(\rv{x}(\tau))$ of the transformed variable $\rv{x}(\tau)$ can be determined via
\begin{equation}
\label{eq:supp-cnf-data-likelyhood}
    \log p(\rv{x}(\tau)) = \log p(\rv{x}(0)) - \int_0^\tau \Tr \left( \frac{\partial g_\rv{\xi}(\rv{x}(t), t)}{\partial \rv{x}(t)}  \right),
\end{equation}
where $\rv{x}(0)$ can be computed from $\rv{x}(\tau)$ by
\begin{equation}
    \rv{x}(0) = \rv{x}(\tau) - \int_0^\tau g_\rv{\xi}(\rv{x}(0), t) \dd t,
\end{equation}
and $\Tr(\cdot)$ is the trace function. Similarly, the training of CNFs amounts to maximizing the data log-likelihood \cref{eq:supp-cnf-data-likelyhood}.

CNFs can be further extended to be conditioned on a vector $\rv{z}$ by using $g_\rv{\xi}(\cdot, t, \rv{z})$ in place of $g_\rv{\xi}(\cdot, t)$. This allows to compute the conditional probability of the data conditioned on $\rv{z}$:
\begin{align}
\label{eq:supp-cnf-data-likelyhood-cond}
    \log p(\rv{x}(\tau) \vert \rv{z}) &= \log p(\rv{x}(0)) - \int_0^\tau \Tr \left( \frac{\partial g_\rv{\xi}(\rv{x}(t), t, \rv{z})}{\partial \rv{x}(t)}  \right);\\
    \rv{x}(0) &= \rv{x}(\tau) - \int_0^\tau g_\rv{\xi}(\rv{x}(0), t, \rv{z}) \dd t.
\end{align}

To sample a CNF,  we first sample an initial point $\rv{x}(0) \sim p(\rv{x}_0)$, then simply follow \cref{eq:supp-cnf-dynamic}.

\section{Training Details}
\label{sec:supp-hyperparameters}
In all experiments we set the latent dimension $D_\rv{z} = 1024$, the number of closest neighbors to consider $k=5$ in \cref{eq:def-rep}, $\sigma_\text{KNN}=0.05$ in \cref{eq:sh-knn-coeffs}, the KKT multiplier (or the coefficient for the high-frequency reconstruction loss) $\eta=5\times10^6$ in \cref{eq:freelbo}, and the maximum degree $L=50$ and $\sigma_\text{Fre}=50$ in \cref{eq:freq-rectifiers}.

We did a grid search to setup these hyperparameter values. Specifically, $\sigma_\text{KNN}$ candidates were set at 0.001, 0.01, 0.05, and 0.1, and $\sigma_\text{Fre}$ at 1, 10, 50, 100. The rationale for a smaller $\sigma_\text{KNN}$ is to limit the influence of distant points in constructing the representative function for a point cloud. Conversely, a larger $\sigma_\text{Fre}$ is preferred to take into account the reconstruction of both high- and low-frequency data. Additionally, $L$ values of 25, 50, 75, and 100 were tested; higher $L$ values enhance frequency domain partition accuracy but increase computational demands. The choice of $L=50$ aligns with common practice in spherical harmonics examples.

We use an Adam optimizer with an initial learning rate of $10^{-3}$ for VAE training and $10^{-5}$ for latent DDPM training with $\beta_1 = 0.9$ and $\beta_2 = 0.999.$ We use a weight decay of $10^{-8}$. During training, the learning rate is decayed by a factor of $10$ whenever the loss plateaus for more than five epochs.

We utilized the Python library \texttt{torch\_harmonics} \cite{bonev2023spherical} for differentiable computation of coefficients as per \cref{eq:spherical-harmonics-coefficients}.

We run all experiments on a machine with a single GPU Nvidia GeForce RTX 4090. Where relevant, all DDPMs are sampled using 1000 time steps.

\section{Further Quantitative Comparisons}
\label{sec:supp-full-table}
We present further quantitative comparisons of different models on point cloud generation task by considering more metrics and more ShapeNet data classes. We employ three evaluative scores for generative models: minimum matching distance (MMD)~\cite{achlioptas2018learning}, coverage (COV)~\cite{achlioptas2018learning}, and 1-nearest neighbor (1-NNA)~\cite{lopez2016revisiting}. Each score can be computed using Chamfer distance (CD) or earth mover distance (EMD), totaling six metrics. While MMD gauges generation fidelity, it remains insensitive to suboptimal samples. Conversely, COV quantifies generation diversity, and 1-NNA measures the distributional similarity between two sets of point clouds, taking both diversity and quality into account. More detailed discussion regarding these metrics can be found in~\cite{yang2019pointflow}.

\Cref{tab:exp-big-table} is the full version of \cref{tab:exp-table} for quantitative comparison of point cloud generation task on three classes of ShapeNet dataset. \Cref{tab:exp-ablation-big} is the full version of \cref{tab:exp-ablation} for FrePolad's ablation study. \Cref{tab:exp-upsampling-compare-big} is the full version of \cref{tab:exp-upsampling-compare} comparing the performance of FrePolad's direct generation of dense point clouds against generation via upsampling from sparce point clouds using various upsampling methods.

\begin{table*}
\centering
{
\begin{tabular}{clcccccc}
\toprule
\multicolumn{1}{c}{\multirow{2}{*}{Class}} & \multicolumn{1}{c}{\multirow{2}{*}{Model}} & \multicolumn{2}{c}{MMD $\downarrow$} & \multicolumn{2}{c}{COV (\%) $\uparrow$} & \multicolumn{2}{c}{1-NNA (\%) $\downarrow$} \\
\cmidrule(lr){3-4} \cmidrule(lr){5-6} \cmidrule(lr){7-8}
\multicolumn{1}{c}{} & \multicolumn{1}{c}{} & \multicolumn{1}{c}{CD} & \multicolumn{1}{c}{EMD} & \multicolumn{1}{c}{CD} & \multicolumn{1}{c}{EMD} & \multicolumn{1}{c}{CD} & \multicolumn{1}{c}{EMD} \\
\midrule
\multirow{12}{*}{Airplane} & Training set & 0.218 & 0.373 & 46.91 & 52.10 & 64.44 & 64.07 \\
\cmidrule{2-8}
 & r-GAN~\cite{achlioptas2018learning} & 0.447 & 2.309 & 30.12 & 14.32 & 98.40 & 96.79 \\
 & 1-GAN (CD)~\cite{achlioptas2018learning} & 0.340 & 0.583 & 38.52 & 21.23 & 87.30 & 93.95 \\
 & 1-GAN (EMD)~\cite{achlioptas2018learning} & 0.397 & 0.417 & 38.27 & 38.52 & 89.49 & 76.91 \\
 & PointFlow~\cite{yang2019pointflow} & 0.224 & 0.390 & 47.90 & 46.41 & 75.68 & 70.74 \\
 & SoftFlow~\cite{kim2020softflow} & 0.231 & 0.375 & 46.91 & 47.90 & 76.05 & 65.80 \\
 & SetVAE~\cite{kim2021setvae} & \textbf{0.200} & 0.367 & 43.70 & 48.40 & 76.54 & 67.65 \\
 & ShapeGF~\cite{cai2020learning} & 0.313 & 0.637 & 45.19 & 40.25 & 81.23 & 80.86 \\
 & DPF-Net~\cite{klokov2020discrete} & 0.264 & 0.409 & 46.17 & 48.89 & 75.18 & 65.55 \\
 & DPC~\cite{luo2021diffusion} & 0.213 & 0.572 & 48.64 & 33.83 & 76.42 & 86.91 \\
 & PVD~\cite{zhou20213d} & 0.224 & 0.370 & \textbf{48.88} & \textbf{52.09} & 73.82 & 64.81 \\
 & LION~\cite{vahdat2022lion} & 0.219 & 0.372 & 47.16 & 49.63 & 67.41 & \textbf{61.23} \\
 & FrePolad (ours) & 0.204 & \textbf{0.353} & 45.16 & 47.80 & \textbf{65.25} & 62.10 \\
 \midrule
\multirow{12}{*}{Chair} & Training set & 2.618 & 1.555 & 53.02 & 51.21 & 51.28 & 54.76 \\
\cmidrule{2-8}
 & r-GAN~\cite{achlioptas2018learning} & 5.151 & 8.312 & 24.27 & 15.13 & 83.69 & 99.70 \\
 & 1-GAN (CD)~\cite{achlioptas2018learning} & 2.589 & 2.007 & 41.99 & 29.31 & 68.58 & 83.84 \\
 & 1-GAN (EMD)~\cite{achlioptas2018learning} & 2.811 & 1.619 & 38.07 & 44.86 & 71.90 & 64.65 \\
 & PointFlow~\cite{yang2019pointflow} & \textbf{2.409} & 1.595 & 42.90 & 50.00 & 62.84 & 60.57 \\
 & SoftFlow~\cite{kim2020softflow} & 2.528 & 1.682 & 41.39 & 47.43 & 59.21 & 60.05 \\
 & SetVAE~\cite{kim2021setvae} & 2.545 & 1.585 & 46.83 & 44.26 & 58.84 & 60.57 \\
 & ShapeGF~\cite{cai2020learning} & 3.724 & 2.394 & 48.34 & 44.26 & 58.01 & 61.25 \\
 & DPF-Net~\cite{klokov2020discrete} & 2.536 & 1.632 & 44.71 & 48.79 & 62.00 & 58.53 \\
 & DPC~\cite{luo2021diffusion} & 2.399 & 2.066 & 44.86 & 35.50 & 60.05 & 74.77 \\
 & PVD~\cite{zhou20213d} & 2.622 & 1.556 & 49.84 & 50.60 & 56.26 & 53.32 \\
 & LION~\cite{vahdat2022lion} & 2.640 & 1.550 & 48.94 & \textbf{52.11} & 53.70 & \textbf{52.34} \\
 & FrePolad (ours) & 2.542 & \textbf{1.532} & \textbf{50.28} & 50.93 & \textbf{52.35} & 53.23 \\
 \midrule
\multirow{12}{*}{Car} & Training set & 0.938 & 0.791 & 50.85 & 55.68 & 51.70 & 50.00 \\
\cmidrule{2-8}
 & r-GAN~\cite{achlioptas2018learning} & 1.446 & 2.133 & 19.03 & 6.539 & 94.46 & 99.01 \\
 & 1-GAN (CD)~\cite{achlioptas2018learning} & 1.532 & 1.226 & 38.92 & 23.58 & 66.49 & 88.78 \\
 & 1-GAN (EMD)~\cite{achlioptas2018learning} & 1.408 & 0.899 & 37.78 & 45.17 & 71.16 & 66.19 \\
 & PointFlow~\cite{yang2019pointflow} & \textbf{0.901} & 0.807 & 46.88 & 50.00 & 58.10 & 56.25 \\
 & SoftFlow~\cite{kim2020softflow} & 1.187 & 0.859 & 42.90 & 44.60 & 64.77 & 60.09 \\
 & SetVAE~\cite{kim2021setvae} & 0.882 & \textbf{0.733} & 49.15 & 46.59 & 59.94 & 59.94 \\
 & ShapeGF~\cite{cai2020learning} & 1.020 & 0.824 & 44.03 & 47.19 & 61.79 & 57.24 \\
 & DPF-Net~\cite{klokov2020discrete} & 1.129 & 0.853 & 45.74 & 49.43 & 62.35 & 54.48 \\
 & DPC~\cite{luo2021diffusion} & 0.902 & 1.140 & 44.03 & 34.94 & 68.89 & 79.97 \\
 & PVD~\cite{zhou20213d} & 1.007 & 0.794 & 41.19 & 50.56 & 54.55 & 53.83 \\
 & LION~\cite{vahdat2022lion} & 0.913 & 0.752 & \textbf{50.00} & \textbf{56.53} & 53.41 & 51.14 \\
 & FrePolad (ours) & 0.904 & 0.782 & 50.14 & 55.23 & \textbf{51.89} & \textbf{50.26} \\
 \bottomrule
\end{tabular}
}
\caption{Quantitative comparison of point cloud generation on three classes of ShapeNet dataset. MMD-CD is multipled by $10^3$ and MMD-EMD is multipled by $10^2$. FrePolad achieves significant improvement over most existing models and obtains state-of-the-art results with simpler and more flexible structure.}
\label{tab:exp-big-table}
\end{table*}

\begin{table*}
\centering
{
\begin{tabular}{clcccccc}
\toprule
\multicolumn{1}{c}{\multirow{2}{*}{Class}} & \multicolumn{1}{c}{\multirow{2}{*}{Model}} & \multicolumn{2}{c}{MMD $\downarrow$} & \multicolumn{2}{c}{COV (\%) $\uparrow$} & \multicolumn{2}{c}{1-NNA (\%) $\downarrow$} \\
\cmidrule(lr){3-4} \cmidrule(lr){5-6} \cmidrule(lr){7-8}
\multicolumn{1}{c}{} & \multicolumn{1}{c}{} & \multicolumn{1}{c}{CD} & \multicolumn{1}{c}{EMD} & \multicolumn{1}{c}{CD} & \multicolumn{1}{c}{EMD} & \multicolumn{1}{c}{CD} & \multicolumn{1}{c}{EMD} \\
\midrule
\multirow{5}{*}{Airplane} & Training set & 0.218 & 0.373 & 46.91 & 52.10 & 64.44 & 64.07 \\
\cmidrule{2-8}
& Po & 0.871 & 0.804 & 23.44 & 35.06 & 85.93 & 85.94 \\
 & FrePo & 0.757 & 0.730 & 26.56 & 40.38 & 80.38 & 80.38 \\
 & Polad & 0.279 & 0.383 & 45.12 & 47.96 & 69.26 & 65.26 \\
 & FrePolad & \textbf{0.204} & \textbf{0.353} & \textbf{45.16} & \textbf{47.80} & \textbf{65.25} & \textbf{62.10} \\
 \midrule
\multirow{5}{*}{Chair} & Training set & 2.618 & 1.555 & 53.02 & 51.21 & 51.28 & 54.76 \\
\cmidrule{2-8}
 & Po & 5.253 & 3.393 & 17.19 & 14.06 & 94.53 & 99.22 \\
 & FrePo & 5.160 & 2.498 & 24.06 & 27.19 & 89.06 & 82.19 \\
 & Polad & 2.721 & 1.642 & 48.23 & 47.73 & 57.35 & 61.52 \\
 & FrePolad & \textbf{2.542} & \textbf{1.532} & \textbf{50.28} & \textbf{50.93} & \textbf{52.35} & \textbf{53.23} \\
 \midrule
\multirow{5}{*}{Car} & Training set & 0.938 & 0.791 & 50.85 & 55.68 & 51.70 & 50.00 \\
\cmidrule{2-8}
 & Po & 1.866 & 11.24 & 18.75 & 9.375 & 100.0 & 100.0 \\
  & FrePo & 1.011 & 2.057 & 27.81 & 28.75 & 89.22 & 84.53 \\
 & Polad & 1.148 & 0.817 & 46.23 & 49.10 & 58.12 & 56.13 \\
 & FrePolad & \textbf{0.904} & \textbf{0.782} & \textbf{50.14} & \textbf{55.23} & \textbf{51.89} & \textbf{50.26} \\
 \bottomrule
\end{tabular}
}
\caption{Ablation study: quantitative comparison of three weakened versions of FrePolad: Polad (without \textbf{f}requency \textbf{re}ctification), FrePo (without \textbf{la}tent \textbf{d}iffusion), and Po (without either). MMD-CD is multipled by $10^3$ and MMD-EMD is multipled by $10^2$. The frequency rectification and the latent DDPM bring significant improvement to FrePolad.}
\label{tab:exp-ablation-big}
\end{table*}

\begin{table*}
\centering
{
\begin{tabular}{clcccccc}
\toprule
\multicolumn{1}{c}{\multirow{2}{*}{Class}} & \multicolumn{1}{c}{\multirow{2}{*}{Model}} & \multicolumn{2}{c}{8192} & \multicolumn{2}{c}{15k} & \multicolumn{2}{c}{100k} \\
\cmidrule(lr){3-4} \cmidrule(lr){5-6} \cmidrule(lr){7-8}
\multicolumn{1}{c}{} & \multicolumn{1}{c}{} & \multicolumn{1}{c}{CD} & \multicolumn{1}{c}{EMD} & \multicolumn{1}{c}{CD} & \multicolumn{1}{c}{EMD} & \multicolumn{1}{c}{CD} & \multicolumn{1}{c}{EMD} \\
\midrule
\multirow{4}{*}{Airplane} & PU-GAN~\cite{li2019pu} & 100 & 99.38 & 100 & 99.06 & 100 & 99.06 \\
 & PU-GCN~\cite{qian2021pu} & 100 & 99.38 & 100 & 99.38 & 100 & 99.38 \\
 & Grad-PU~\cite{he2023grad} & 100 & 99.38 & 100 & 99.38 & 100 & 99.38 \\
 \cmidrule{2-8}
 & FrePolad (ours) & \textbf{65.31} & \textbf{63.88} & \textbf{66.87} & \textbf{57.81} & \textbf{66.25} & \textbf{60.25} \\
 \midrule
\multirow{4}{*}{Chair} & PU-GAN~\cite{li2019pu} & 100 & 100 & 100 & 100 & 100 & 100 \\
 & PU-GCN~\cite{qian2021pu} & 100 & 100 & 100 & 100 & 100 & 100 \\
 & Grad-PU~\cite{he2023grad} & 100 & 100 & 100 & 100 & 100 & 100 \\
 \cmidrule{2-8}
 & FrePolad (ours) & \textbf{52.69} & \textbf{50.63} & \textbf{53.63} & \textbf{50.00} & \textbf{52.69} & \textbf{51.56} \\
 \midrule
\multirow{4}{*}{Car} & PU-GAN~\cite{li2019pu} & 99.06 & 98.75 & 99.06 & 98.75 & 99.06 & 98.13 \\
 & PU-GCN~\cite{qian2021pu} & 99.38 & 99.06 & 99.38 & 99.06 & 99.38 & 99.06 \\
 & Grad-PU~\cite{he2023grad} & 99.38 & 98.75 & 99.38 & 98.75 & 99.38 & 98.75 \\
 \cmidrule{2-8}
 & FrePolad (ours) & \textbf{52.06} & \textbf{54.25} & \textbf{52.06} & \textbf{51.75} & \textbf{51.13} & \textbf{51.50} \\
 \bottomrule
\end{tabular}
}
\caption{1-NNA scores ($\downarrow$) of the generated dense point clouds with different cardinalities. FrePolad directly generates dense point clouds. The first three methods upsample sparse ground truth ones. FrePolad profoundly outperforms all upsampling methods, demonstrating the importance of the ability to directly generate dense point clouds.}
\label{tab:exp-upsampling-compare-big}
\end{table*}

\end{document}